\documentclass[a4paper, 10 pt, conference]{ieeeconf}      
\IEEEoverridecommandlockouts 
\usepackage{graphicx}
\usepackage{amssymb}
\usepackage{algorithm2e}
\usepackage{amsmath}
\usepackage{array}
\usepackage{color}
\usepackage{soul}
\usepackage{bm}
\usepackage{times}
\usepackage[table]{xcolor}
\usepackage{multirow}
\usepackage{xfrac}
\let\labelindent\relax

\usepackage{enumitem}
\usepackage{makecell}
\usepackage{cite}
\usepackage{fontawesome}

\xdefinecolor{gray95}{gray}{0.65}
\xdefinecolor{gray92}{gray}{0.9}
\xdefinecolor{gray25}{gray}{0.8}
\newcolumntype{L}[1]{>{\raggedright\let\newline\\\arraybackslash\hspace{0pt}}m{#1}}
\newcolumntype{C}[1]{>{\centering\let\newline\\\arraybackslash\hspace{0pt}}m{#1}}
\newcolumntype{R}[1]{>{\raggedleft\let\newline\\\arraybackslash\hspace{0pt}}m{#1}}

\makeatletter
\def\endthebibliography{%
	\def\@noitemerr{\@latex@warning{Empty `thebibliography' environment}}%
	\endlist
}
\makeatother

\title{\LARGE \bf
Transfer Learning and Online Learning for Traffic Forecasting under Different Data Availability Conditions: Alternatives and Pitfalls
}

\author{Eric L. Manibardo$^{1,2}$,  Ibai La\~ na$^{1}$, and Javier Del Ser$^{1,2}$
\thanks{$^{1}$ Eric L. Manibardo, Ibai La\~na, and Javier Del Ser are with TECNALIA, Basque Research and Technology Alliance (BRTA), 48160 Derio, Bizkaia, Spain. Contact email: eric.lopez@tecnalia.com}%
\thanks{$^{2}$Eric L. Manibardo and Javier Del Ser are with the University of the Basque Country (UPV/EHU), 48013 Bilbao, Bizkaia, Spain.}%
}

\begin{document}

\maketitle
\thispagestyle{empty}
\pagestyle{empty}

\begin{abstract}
This work aims at unveiling the potential of Transfer Learning (TL) for developing a traffic flow forecasting model in scenarios of absent data. Knowledge transfer from high-quality predictive models becomes feasible under the TL paradigm, enabling the generation of new proper models with few data. In order to explore this capability, we identify three different levels of data absent scenarios, where TL techniques are applied among  Deep Learning (DL) methods for traffic forecasting. Then, traditional batch learning is compared against TL based models using real traffic flow data, collected by deployed loops managed by the City Council of Madrid (Spain). In addition, we apply Online Learning (OL) techniques, where model receives an update after each prediction, in order to adapt to traffic flow trend changes and incrementally learn from new incoming traffic data. The obtained experimental results shed light on the advantages of transfer and online learning for traffic flow forecasting, and draw practical insights on their interplay with the amount of available training data at the location of interest.
\end{abstract}


\section{Introduction}
One of the most sought-after applications by Intelligent Transportation Systems (ITS) is short-term traffic forecasting \cite{vlahogianni2014short}. Possibilities rooted on this paradigm are manifold. For instance, effectively predicting near-future traffic congestion states would grant more flexibility (whenever necessary) to traffic management operators for the application of proper countermeasures. Such a powerful tool is under the spotlight of most major cities in the world, where other procedures such as the increase of infrastructures' capacity or the promotion of other transportation alternatives do not provide enough assistance to dwindle congestion situations \cite{lana2018road}. However, the lack of data to effectively model traffic profiles at a certain location is not an unusual scenario. Indeed, there lies the motivation of this work: to determine the most suitable modeling techniques to implement a forecasting model for contexts of varying data availability, considering for this purpose the actual state of the art. 

When it comes to modeling, Deep Learning (DL) networks have been successfully applied to traffic forecasting in recent years, taking advantage of the increasing amount of historical data. Neural networks within the DL family is composed by several hidden layers with multiple nodes, that allow capturing complex non-linear representations of data. Inputs of first layer are multiplied by different weight values of its nodes, obtaining an output vector that serves as input of next layer and so on. Weight values are updated after calculating the error produced between a training sample and the predicted instance by the network. In the case of traffic forecasting, early works used only neural layers to conform a stacked auto-encoder \cite{lv2014traffic, yang2016optimized}. Later studies started to explore the capabilities of recurrent neural networks, in particular Long Short-Term Memory (LSTM) cells \cite{zhao2017lstm}, often mixed with convolutional layers in order to extract high level features \cite{yu2017spatiotemporal, yao2018deep,yao2019revisiting}, claiming that spatio-temporal correlation was giving valuable information to solve the prediction task. The so-called convolutional recurrent neural network, which combines convolutional and recurrent layers, has been set as the prevailing architecture for time series modeling. In this work we will embrace this hybrid modeling architecture.

In tasks undergoing a severe scarcity of annotated data, techniques to exploit previously acquired knowledge to solve new yet similar problems have emerged and grouped under the Transfer Learning (TL) paradigm. Several comprehensive overviews have gravitated on TL methods and their applications \cite{pan2009survey,lu2015transfer,weiss2016survey}. Interestingly, DL based techniques have lately dominated this research arena, which permit to minimize the required amount of training data, to increase learning rate and, sometimes, even to achieve an improved predictive performance. Typically, when developing a predictive modeling for a particular task, labeled data is required. After annotating each example with its proper tag, which is the value to predict, the dataset is split into train/test data. Since the data source is the same, both data collections have identical data distribution. However, as time passes, the incoming instances to be predicted (i.e. current test data) may change their data distribution. At this moment, prediction model can not be used and must be rebuilt from scratch using new training data, whose collection can be expensive and sometimes even impossible in practice. In this context, TL provides algorithmic means to transfer already acquired knowledge from a source domain to a target domain, the latter possibly with a different yet related data distribution. The work of \textit{Hu et al.} \cite{hu2016transfer} offers a nice case study about the capabilities of TL, where they need to predict wind speed at newly-built farms. Needless to say that sufficient historical data is not available for training an accurate model, so they propose to transfer the predictive knowledge captured over older wind farms that have long-term records. The aforementioned problems also arise within the transportation domain, where the recent state of the art \cite{chen2019mmse} shows a similar approach to the one presented in this paper. The lack of information at early stages implies an hindrance for the fast deployment of predictive models over the road network at multiple points. Our postulated hypothesis is that TL techniques may allow lowering the amount of historical data needed in new locations, thereby accelerating model learning and ultimately, its deployment.  

Lastly, we explore the benefits that TL may produce under an online learning (OL) scenario, by which models update their captured knowledge as per the supervision of newly arriving test instances over time. Close to the paper of \textit{Zhao et al.} \cite{zhao2014online}, OL provides small updates to the transferred model, adjusting the traffic data predictor to the actual trend. In this way, network weights would not be initialized at random, but they would rather start from suitably transferred weight values, which would be adapted on-line to the behaviour of target location. Our thoughts are that this methodology could provide faster convergence to an optimal solution.

To sum up, this work focuses on the capabilities of TL and OL in the filed of traffic data forecasting. Specifically, we aim for TL to accelerate deployment phase by minimizing training data needs, due to prior acquired knowledge from transferred models. On the other hand, OL adapts the learned knowledge of the model to the prevailing characteristics of traffic data, improving the model performance over time. The main contribution of this paper revolves around the benefits that could bring different learning strategies relying on OL and TL, depending on the amount of available data.

The rest of the manuscript is structured as follows: Section \ref{sec:methods} details the data utilized for our study, and defines the different learning strategies under consideration. Next, the experimental setup is described at Section \ref{sec:exp}, whereas Section \ref{sec:results} gathers and discusses the obtained results. Finally, Section \ref{sec:conc} ends the paper by giving insights on which learning techniques to implement on scenarios of different data scarcity levels, along with future research lines derived from our findings.

\section{Materials and Methods} \label{sec:methods}
\begin{figure}[h!]
\centering
\includegraphics[width=0.75\columnwidth]{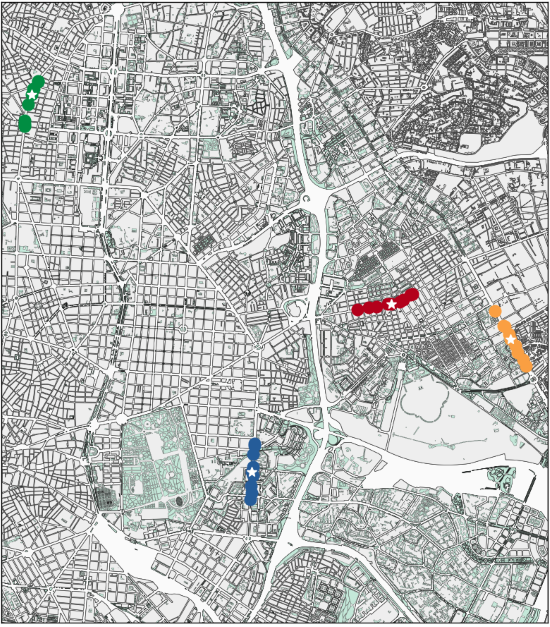}
\caption{Location of the selected loops around the M-30 surrounding area. Colored markers are kept throughout the study: Alcalá (red), Bravo Murillo (green), Esquerdo (blue) and García Noblejas (orange). The white star markers \faStarO \hspace{0.1cm} denote the loops where predictions are made.}
\label{fig:MAP}
\end{figure}

Data for this research work have been collected from a public repository maintained by the City Council of Madrid (Spain) \cite{DatosMadrid}. We have chosen traffic flow data, posted in the form of 15-minute aggregated periods during 2017 and 2018 years \cite{lana2016understanding}. Data was recorded by sensors located in urban arterials, placed close to the main belt of the city, namely, the so-called M-30 highway. Then, four locations have been selected (shown in Figure \ref{fig:MAP}) towards considering different traffic profiles based on their number of lanes and proximity to Madrid center: Alcalá, Bravo Murillo, Doctor Esquerdo and García Noblejas streets.

Usually, bottleneck states are propagated downstream, in opposite direction to the traffic flow. Even so, there are some conditions where this transmission occurs upwards along the road \cite{cassidy1999some}. Under this premise, we have designed the following scheme: if the flow value of a certain loop A is to be predicted at slot $t$, features are defined as $\{t-5, \dots ,t-1\}$ instant flow values, recorded at the next four and previous four loops placed in the vicinity of A, along with $\{t-5, \dots ,t-1\}$ slot flow values from loop A itself. Consequently, a set of $45$ historical input values are given to the model in order to predict the next flow value in a 15-minute interval. This modeling choice assumes that previous flow values from surroundings and target location itself, contain enough predictive information to build a proper short-term forecasting model. 

\begin{figure*}[ht!]
	\centering
	\includegraphics[width=1.85\columnwidth]{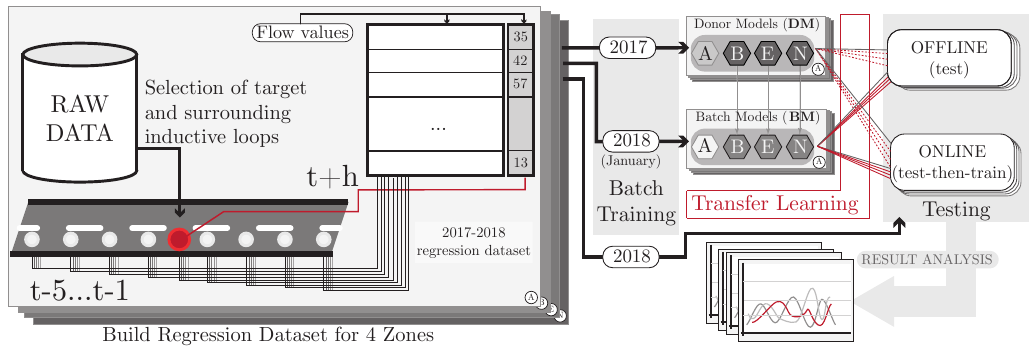}
	\caption{Experimentation workflow. Datasets for each year are built based on traffic flow values collected from selected loops of 4 different roads (A, B, E, and N, respectively). The dataset corresponding to 2017 contains information used for training donor models (DM), while the first month of 2018 is used to train a model from scratch and to re-train DM copies (only those from different placements with regard to target location). Then, all models are tested under offline and online settings, producing the results analyzed and discussed in Section \ref{sec:results}.}
	\label{fig:workflow}
	\vspace{-3mm}
\end{figure*}

\subsection{Deep Learning Architecture}

Given that the focus of our work is placed on knowledge transfer (TL) and updating (OL) strategies, we devise a general DL modeling architecture that performs properly for all considered locations. Nowadays, modeling trends in traffic forecasting are arguably monopolized by convolutional recurrent neural networks \cite{yu2017spatiotemporal, yao2018deep, yao2019revisiting}. We therefore opt for a similar approach. Our architecture receives an input of 9 vectors containing 5 values each, formed by $\{t-5, \ldots ,t-1\}$ traffic flow values of each $4+1+4=9$ available loops for the road under analysis. The total of 45 feature values goes through 50 one-dimensional convolutional filters, of size two coefficients. This process allows extracting high-dimensional features from input vectors. Then, the convolutional layer output is fed to a stateful LSTM layer \cite{zhao2017lstm} of 75 memory cells, endowed with the role of discovering long temporal dependencies over time. Finally, a dense layer of 50 neurons selects the most significant output values, making prediction of the future traffic flow level. This DL architecture is applied to model every traffic profile analyzed in this study. 

\subsection{Transfer Learning Technique} 

With the aim to transfer knowledge from one model to another, we must first select a related predictive modeling problem. The related task should have enough data to develop a skillful model, where strong concepts relating input to output data are learned. If the performance is below that of a na\"ive model, no valuable knowledge can be transmitted, so it is important to overcome this barrier. Next, the already prepared model on the source task is used as starting point for the model on the target task. In DL, this is done by copying the adjusted weights to the network of the target task.

\subsection{Online Setting}

To the best of our knowledge, a few studies can be found mixing traffic data with OL \cite{niu2015online, chen2016long}, possibly due to the relatively large time gap between arrival samples with respect to more traditional OL tasks, where consecutive instances turn up in typically less than one minute  \cite{seufert2019stream}. However, upon the availability of streaming data, small updates can be given to the model, securing the improvement of the already learned knowledge. Moreover, we compare the performance evolution of two models which depart from the same acquired knowledge, under online and offline settings, maintaining the last one unaltered during the test phase.

\subsection{Regression Metric}

In this study we have selected the coefficient of determination $R^2$ as regression metric. This coefficient expresses the quality of the forecasting model, as it measures the variance between real and predicted values. We compute $R^2$ for each slot $t$ averaged over a full week sized window, in order to show performance changes originated by traffic flow drifts. Having samples every 15 minutes provides 96 daily samples, which results in windows that comprehend 672 slots:
\begin{equation} \label{eq:r2}
R^{2} \doteq 1 - \left(\sum\limits_{t=n-672}^{n} {(o_t - \hat o_t)}^2 \bigg/ \sum\limits_{t=n-672}^{n} {(o_t - \bar {o_t} })^2\right),
\end{equation}
where $o_t$ denotes the real observed value at slot $t$, $\bar{o_t}$ its average, and $\hat o_t$ the predicted one.

\section{Experimental Setup} \label{sec:exp}

As mentioned before, this work pursues to identify the most appropriate modeling strategy when predicting traffic flow at a specific placement, according to the amount of available data. The forecasting horizon is set to 1 slot (i.e .15-minute interval). In addition, we assume the disposal of historical traffic flow data for another three locations of the same city, so that the knowledge captured by a forecasting model in these locations will be exported to the model developed for the location of interest, whose release date is set to 01 January 2018. With this methodology in mind, we identify three Possible Scenarios (PS):
\begin{itemize}[leftmargin=*]
    \item PS1: Only historical data at other locations is available. This means that no data is available for the location of interest, so the only option is to develop a model in another road, and transfer it to the target placement.
    \item PS2: Historical data is available at other locations, along with a few weeks of historical data at the target site. Data at target location should be collected, for example, by deploying surface loops. Then, the knowledge contributed by models learned from data of other roads (donor models) can be specialized by learning from this temporal data collected at the target placement. It is possible also to train a model from scratch, only using temporal gathered data.
    \item PS3: Historical data is available at target site. Traditional batch learning model at target location can be developed.
\end{itemize}

In order to discover the right setting for every PS, we have designed the experimental setup shown at Figure \ref{fig:workflow}. The steps of this process are detailed next: firstly, we develop four regression models, one per selected road described in Section \ref{sec:methods}. The goal is to provide high performance forecasting models, so full 2017 year historical data is fed to DL network for training. This way, the network is trained on examples of every day along the year, learning both usual traffic profiles from normal working periods, and special events such as Christmas or Easter holidays, where traffic profile changes no matter the weekday. Therefore, we train by using one day sized batches for 10000 epochs without shuffling, to refresh the model as per the evolution of traffic profile sequentially. The same DL architecture is used for all experiments. 

At PS1 scenario, data is not available at target scenario, so we can only exploit information from other roads. Thus, we transfer pre-trained models from those roads to the model of the target scenario, via TL. Then, test phase is performed from January first, until the last day of 2018, covering all existing days. Now, if we have few weeks of historical data (we set this for the full month of January), collected by temporal surface loops like in PS2 scenario, two new options emerge. The first one is to retrain transferred models by using target location data. This way, we allow models to adapt from their original concepts to the actual one, throughout 2000 epoch. We set less epochs due to lower amount of training data. The second option consists of directly train batch-wise a new model, by using January 2018 data at target road. The main drawback of both strategies is that model release date would be delayed until February. Moreover, our hypothesis is that examples that conform training data, may have noticeable impact over model behaviour when facing the prediction of special-event days. In fact, at Spain the traffic profile of the first week of the year is quite unique, because of New Year's and Epiphany day (national festivity at January 6), producing flow peaks at certain hours, when people start or end they holiday trips. Lastly, the PS3 scenario allows to prepare a model drawn from 2017 data at target location. All the knowledge collected in a whole year should produce the highest quality possible model, because the dataset actually has examples for all different special events at target placement.

In addition to the previously explained tests, an online version of each model was also added to the comparison study, in order to assess how much performance can be improved with respect its offline counterpart. Until now, the only source of knowledge came from TL and batch learning. After that, the model only predicts the next traffic flow value. In contrast, OL imprints small updates to the model over time, by using the incoming real value samples. Under this paradigm, we test the model by comparing predicted value to real and then, we use actual value as learning example for 1 epoch. It is important to perform only one back-propagated gradient update per tested sample; otherwise, the adaptation of the weights of the model could become overly biased to just one instance. 

\section{Results and Discussion} \label{sec:results}

\begin{figure*}[ht!]
	\centering
	\includegraphics[width=2\columnwidth]{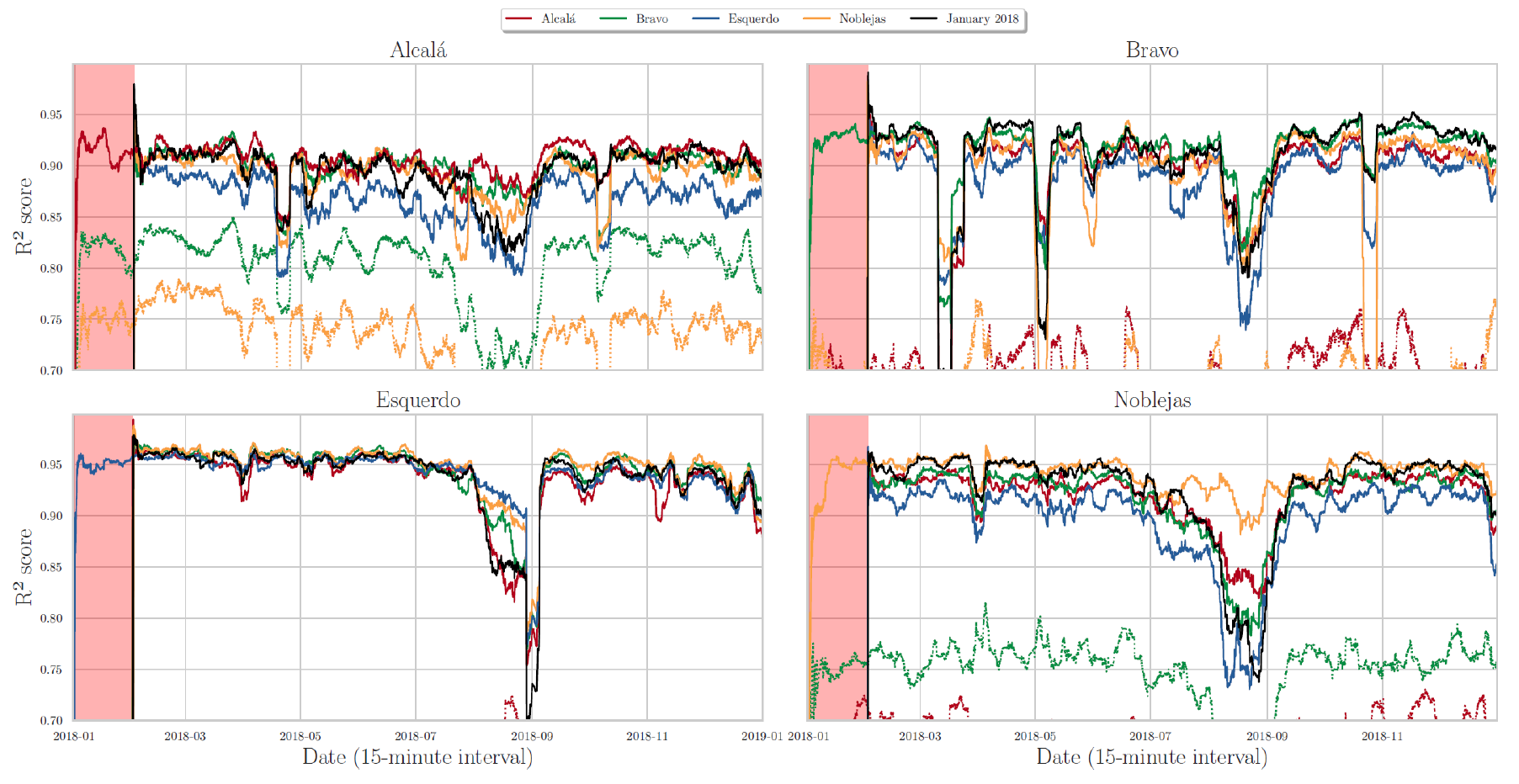}
	\caption{$R^2$ score evolution comparison for offline settings. Colored identifiers denote historical data placement from 2017 year. In the cases where target road and data source location do not match, TL techniques have been applied. The remaining colored line identifies model trained with 2017 full year target location data. In particular, dotted lines means that transferred models were implemented directly at the release date, while continuous lines pinpoint transferred models that were also trained with January 2018 data (red area). These last cases, along with black line which spots a model trained only with January 2018 data from target location, were implemented one month after the release date.}
	\label{fig:offline}
\end{figure*}

The discussion begins by commenting on Figure \ref{fig:offline}, which depicts the $R^2$ score comparison between different offline approaches. As expected, the best performance case correspond to PS3 scenario, where historical data at target location is available (for each subplot at Figure \ref{fig:offline}, the line that matches name with target location). The model learns from different traffic flow events such as Easter or summer holidays, along with regular days, always from the target road. Therefore, this framework offers the most favorable conditions, positioning itself as a performance baseline.

\begin{figure*}[ht!]
	\centering
	\includegraphics[width=2\columnwidth]{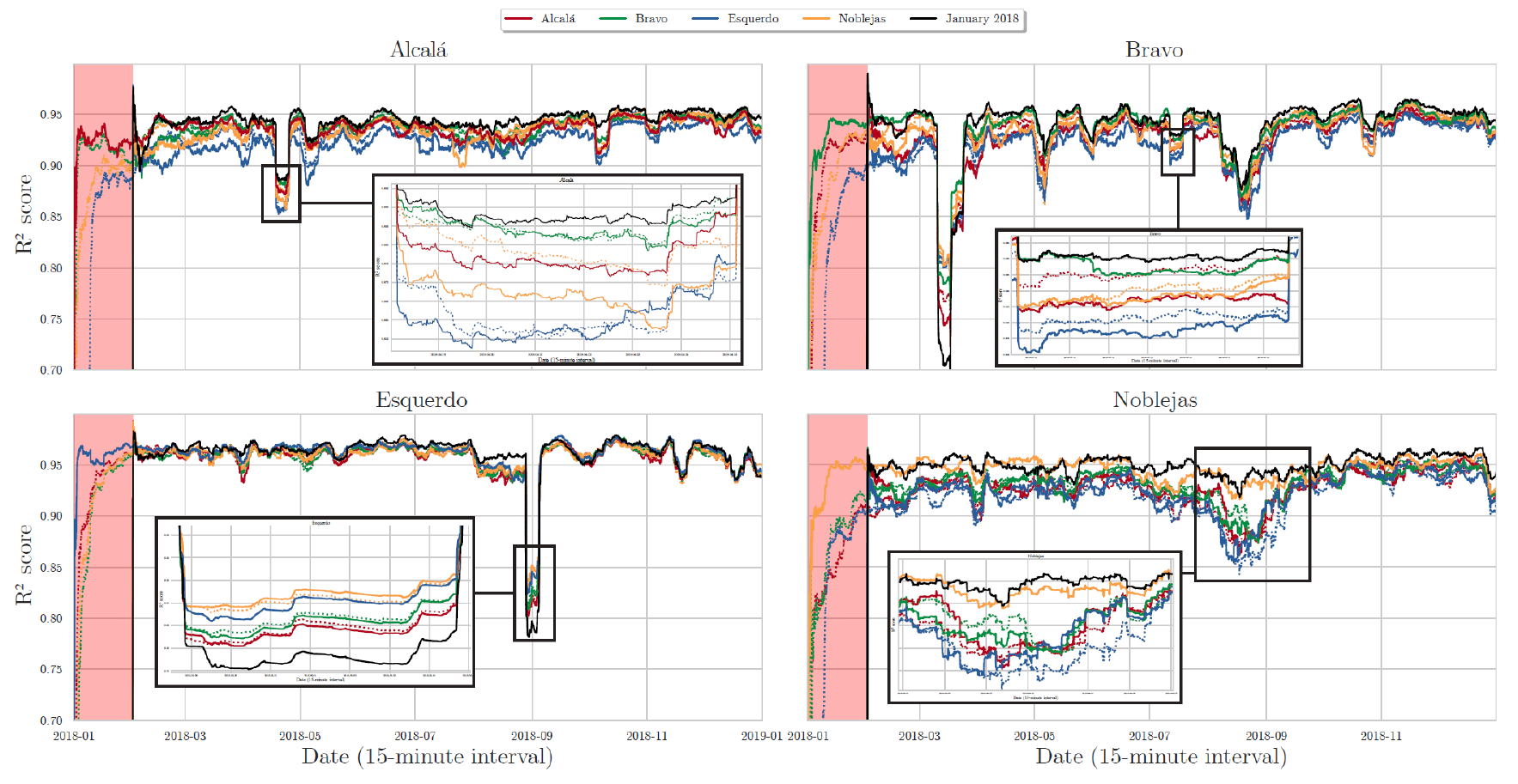}
	\caption{$R^2$ score evolution comparison for online settings, following the color convention of Figure \ref{fig:offline}. Boxes show in detail performance at certain areas.}
	\label{fig:online}
\end{figure*}

Then, the transferred models are analyzed: those which were deployed at release date and the ones which were also trained under PS2 scenario, using temporal loop gathered data. If we focus at dotted lines, that represent the PS1 scenario, we can observe that different transferred models elicit distinct behaviours. For example, Bravo Murillo road based transferred model works quite well for Alcalá and García Noblejas datasets, bearing in mind that no updates were given to the model. However, this approach does not work in the opposite way, where Alcalá and García Noblejas based models perform worse at Bravo Murillo road test. Our hypothesis is that during 2017 year, Bravo Murillo road experienced some events that have had notable impact over transferred model development, allowing to be more prepared to forecast in the course of the test. On its part, Dr. Esquerdo data based transferred model does not perform well over the rest of locations (it obtained negative results, so lines are out of the chart). If we plot traffic historical data, Dr. Esquerdo has larger car flow peaks when compared to other placements. The abruptness of these spikes makes harder to predict when testing transferred models at Dr. Esquerdo and also when transferring Dr. Esquerdo based model to other locations. Consequently, we decided to magnify charts in order to better inspect such details, as blue dotted lines do not offer clear insights.

On the other hand, after training transferred models over January data at target location, all of them improved greatly up to 90\% $R^2$ score. Intuitively, the better their dotted counterparts are, the greater performance is obtained due to the exclusive knowledge learned from other locations. However, if we compare them to the black line, that identifies the behaviour of a model trained exclusively with first month of 2018, under PS2 scenario, we obtain some interesting insights. During regular traffic periods, the model represented by black line performs slightly better, as model has only seen examples from target context. In contrast, when special events occur, like last week of August, when people return from holidays, performance is contingent upon specific knowledge of each model. During the mentioned week, there is always a transferred model that exceed black line. With this, we want to express the importance of showing unique events to model, providing it with the necessary knowledge to deal with the arrival of other exceptional episodes.

Now we focus at OL, under the assumption that there are hardware resources capable of buffering an incoming data record and using it as an instance to update the model and keep it adapted to the prevailing traffic flow patterns. The performance of the models shown at Figure \ref{fig:offline} are now displayed for the online setting at Figure \ref{fig:online}. It is remarkable how dotted lines, which represent transferred models performance at PS1 scenario, improve greatly compared to their offline counterparts, catching up other models, and sometimes even performing better than their PS2 equivalents (transferred models that were re-trained using temporal gathered data). This may be due to specialization of PS2 transferred models over January, against dotted lines that were tested through this month, improving by online updates. This effect is shown at the black boxes of Figure \ref{fig:online}, that offer an expanded view. Furthermore, we remark that PS1 based models were implemented at release date, a month earlier than PS2 based models.

In general, an increase of $R^2$ score was achieved respect offline setting but, what is more important, performance at special events, where offline setting struggles, was greatly improved. Reinforcement offered by OL procures more valuable knowledge for all models, letting them to actually presence additional special events. Models represented by black lines demonstrate remarkable degree of improvement, being the only ones that were trained with such a small dataset. In this case, OL enhances significantly the acquired knowledge, because models have only learned, as singular traffic events, the first week of January. Finally, the model developed under the PS3 scenario prevails as the most reliable one, keeping high performance levels at routine days and showing acceptable predictions during special events, where in some cases models represented by black lines commit more errors.

\section{Conclusions and Future Work} \label{sec:conc}

On a closing note, we summarize the main modeling insights and the best strategy for every PS. First, when OL is not an option (e.g. due to hardware limitations), results have shown that the deployment of temporal loops to collect enough initial data to seed a model can be a good practical choice. One month resulted to suffice as long as it contains both normal periods and special events. In most countries, a large fraction of the population usually implement their holidays in August, making this month non-representative of the traffic behaviour in general terms. Having historical data of representative periods allows training a model that will perform better and without the drawbacks derived from storing historical data at every target location. If PS2 scenario is not an option, and only PS1 is feasible, it is possible to achieve reasonable results via TL, but the source location must have similar traffic flow patterns. Lastly, for the most benign paradigm PS3, we would like to underscore that no major improvements were achieved with regard to PS2. 

However, insights change when OL is considered, rendering similar performance for all PS. Although PS3 based models still offer good and stable performance levels, we advocate for models developed by using PS1 scenario available data. In contrast to PS2 based models, PS1 permits to deploy the model at the fixed release date, as batch training using temporal loop gathered data does not reduce prediction errors when OL is available. Thus, unless historical data at target location is available, which still produces the best of the forecasting accuracies, we propose directly transferred models along with OL updates as the overall best approach. The trade-off between fast implementation, high performance and low target road data necessity of PS1 scenario, makes it the most versatile option. This work have therefore evinced the capabilities of TL, achieving high quality traffic flow forecasting models from an already trained model.  

Future research will steer the focus to placements with more diverse traffic data profiles, from highways comprising several lanes to small village roads, in order to confirm whether the conclusions issued by this study can be extrapolated to every traffic forecasting scenario. In addition, attention will be also paid to pedestrian flow forecasting. Since major cities are pushing forward pedestrian infrastructure towards minimizing carbon emissions, pedestrian flow management will become relevant in the near future. In this regard, accurate forecasts of pedestrian flows will be a key towards safer, more enjoyable and sustainable urban environments.

\section*{Acknowledgments}

The authors thank the Basque Government for its support through the EMAITEK and ELKARTEK funding programs. Eric L. Manibardo receives funding support from the Basque Government through its BIKAINTEK PhD support program (grant no. 48AFW22019-00002). Javier Del Ser receives support from the Consolidated Research Group MATHMODE (IT1294-19) granted by the Department of Education of the Basque Government.

\bibliographystyle{IEEEtran}
\bibliography{biblio.bib}

\end{document}